\documentclass[letterpaper]{article}
\usepackage{xcolor}
\usepackage{graphicx}
\usepackage{url}
\usepackage{flairs}
\usepackage{times}
\usepackage{helvet}
\usepackage{courier}
\begin{document}
\title{Beyond Coefficients: Forecast-Necessity Testing for Interpretable Causal Discovery in Nonlinear Time-Series Models}
\author{Valentina Kuskova, Dmitry Zaytsev\\
Lucy Family Institute for Data \& Society\\
University of Notre Dame\\
Notre Dame, IN, USA\\
\And Michael Coppedge\\
Department of Political Science\\
University of Notre Dame\\
Notre Dame, IN, USA\\
}
\maketitle
\begin{abstract}
\begin{quote}
Nonlinear machine-learning models are increasingly used to discover causal relationships in time-series data, yet the interpretation of their outputs remains poorly understood. In particular, causal scores produced by regularized neural autoregressive models are often treated as analogues of regression coefficients, leading to misleading claims of statistical significance.

In this paper, we argue that causal relevance in nonlinear time-series models should be evaluated through forecast necessity rather than coefficient magnitude, and we present a practical evaluation procedure for doing so. We present an interpretable evaluation framework based on systematic edge ablation and forecast comparison, which tests whether a candidate causal relationship is required for accurate prediction. Using Neural Additive Vector Autoregression as a case study model, we apply this framework to a real-world case study of democratic development, modeled as a multivariate time series of panel data - democracy indicators across 139 countries. We show that relationships with similar causal scores can differ dramatically in their predictive necessity due to redundancy, temporal persistence, and regime-specific effects. 

Our results demonstrate how forecast-necessity testing supports more reliable causal reasoning in applied AI systems and provides practical guidance for interpreting nonlinear time-series models in high-stakes domains. 
\end{quote}
\end{abstract}

\section{Introduction}
Neural autoregressive models are increasingly used to discover and summarize causal relationships in time-series data (Geiger et al., 2015; Bussmann et al., 2021). By estimating directed dependencies among variables, these models produce causal scores that are often interpreted as measures of causal importance (Montagna et al., 2023). In practice, however, such scores are frequently treated as analogues of regression coefficients or test statistics, despite being derived from nonlinear, regularized, and highly dependent systems (Tramontano et al., 2024).

This practice leads to systematic reasoning errors (Wan et al., 2025). In nonlinear time-series models, large causal scores may reflect temporal persistence, redundancy among predictors, or regime-specific effects rather than genuine causal relevance (Troster and Wied, 2021). Conversely, relationships that are critical for accurate forecasting may exhibit modest average contributions and therefore appear unimportant under magnitude-based interpretations. As a result, users may draw misleading conclusions about which relationships matter and why (Wan et al., 2025).

In this paper, we argue that causal relevance in nonlinear time-series models should be evaluated through forecast necessity rather than coefficient magnitude. Intuitively, a relationship is causally important if it is required for accurate prediction: removing it should degrade out-of-sample performance. This notion aligns more closely with counterfactual reasoning and avoids conflating frequent or large contributions with indispensable ones. Critically, this is a practical evaluation procedure, not a redefinition of causality: it operationalizes a Granger-style predictive-dependence criterion for nonlinear, regularized settings where coefficient-based tests are unavailable.

We introduce a practical evaluation framework based on systematic edge ablation (Job et al., 2025) and forecast comparison. For each candidate causal relationship, we compare the predictive accuracy of the full model to that of a restricted model in which the relationship is removed. Forecast degradation is assessed using standard forecast comparison tests, yielding a clear operational criterion for necessity.

We demonstrate this framework using Neural Additive Vector Autoregression (NAVAR) (Bussmann et al., 2021) as a representative nonlinear time-series model. We apply the method to a real-world case study of democratic development, modeled as a multivariate time series of democracy indicators across countries. The goal of this case study is not to adjudicate political theories, but to illustrate how different notions of causal relevance lead to qualitatively different interpretations in a complex interpretable system.

Our contributions are threefold. First, we formalize forecast necessity as an interpretable criterion for causal relevance in nonlinear time-series models. Second, we propose a general evaluation framework based on edge ablation and forecast comparison. Finally, we show that magnitude-based causal scores and forecast necessity can diverge substantially in practice, with important implications for causal reasoning.

\section{Background and Related Work}
\subsection{Causal Discovery in Time Series}

Classical approaches to causal discovery in time series, such as Granger causality (Zhou et al., 2024; Zhang et al., 2024) and vector autoregression (Li and Genton, 2009), rely on linear models and coefficient-based inference. Extensions to nonlinear settings include kernel methods (Lim et al., 2015), state-space models (Aoki, 2013), and neural autoregressive architectures (Taskaya-Temizel and Casey, 2005). These models offer greater expressive power but complicate interpretation, as learned parameters no longer correspond directly to marginal effects or testable hypotheses.

Neural autoregressive models (Chen et al., 2025) typically summarize learned dependencies using heuristic measures such as contribution magnitudes, gradient norms, or variability scores. While useful for visualization, these quantities lack a clear inferential interpretation and are often mistaken for indicators of statistical or causal significance. The gap this paper addresses is precisely this: existing magnitude-based summaries tell practitioners \textit{how much} a variable contributes on average, but not \textit{whether} that contribution is required for prediction. Forecast-necessity testing fills this gap by providing a behavioral, testable criterion grounded in the Granger causality tradition, extended to nonlinear settings.

\subsection{Interpretability and Explanation in Machine Learning}

A large literature addresses the interpretability of complex models through post-hoc explanations, including feature attribution methods such as Shapley Additive exPlanations (SHAP) (Albini et al., 2022) and integrated gradients (Kwon, Koo, and Cho, 2021). These tools explain how predictions are formed locally, but they do not determine whether a feature is necessary for predictive performance. Consequently, explanation alone cannot resolve questions of causal relevance or indispensability.

Recent work emphasizes the need for evaluation-based interpretability (Kruschel et al., 2025), in which explanations are tied to observable changes in model behavior. Our approach follows this line by evaluating causal claims through their impact on predictive accuracy.

\subsection{Forecast Comparison as Evidence}

Forecast comparison methods (Mehdiyev et al., 2016) are widely used in econometrics and time-series analysis to assess whether one model predicts better than another. Tests such as the Diebold-Mariano test (Diebold, 2015) provide a principled way to evaluate differences in forecast accuracy while accounting for temporal dependence.

We repurpose these tools to evaluate causal relationships rather than entire models. By comparing forecasts before and after removing a candidate relationship, we obtain a direct test of whether that relationship is necessary for prediction.\\

\section{Forecast Necessity as Causal Relevance}
\subsection{Definition} We define a causal relationship as forecast-necessary if removing it from a predictive model results in a statistically meaningful degradation in out-of-sample forecasting performance. Necessity is therefore a counterfactual property (Kment, 2006): it concerns what the model can no longer predict once the relationship is absent. This definition deliberately avoids claims about structural causality. Instead, it operationalizes causal relevance as a property of predictive dependence that can be empirically tested (Nichols and Cooch, 2025), in the spirit of Granger causality extended to nonlinear settings.

\subsection{Why Magnitude Is Insufficient} Magnitude-based causal scores conflate several distinct phenomena. Large scores may arise from strong temporal persistence, correlated predictors, or effects that are large but infrequent. In such cases, a relationship may appear important despite being substitutable by other variables. Conversely, small but consistent effects may be critical for prediction and yet be overlooked. As a result, magnitude alone does not answer the question practitioners care about: Is this relationship required for the model to work?

\subsection{Implications for Causal Reasoning} Evaluating forecast necessity shifts causal reasoning from coefficient inspection to behavioral assessment. It encourages interpretations grounded in observable consequences - loss of predictive ability - rather than abstract parameter values. This shift is particularly important in nonlinear, high-stakes domains, where misinterpreting causal outputs can lead to unwarranted confidence and poor decision-making. In the remainder of the paper, we operationalize this concept through a simple and general evaluation framework and demonstrate its value in a real-world, interpretable system.

\section{Methodology}
\subsection{Problem Setup}

Let ${\bf y}_t = (y_{1t}, \dots, y_{Nt})^\top$ denote an $N$-dimensional multivariate time series observed over $t=1,\ldots,T$. For a fixed lag order $p$, we consider predictive models of the form
\[
{\bf y}_t = f({\bf y}_{t-1}, \dots, {\bf y}_{t-p}) + {\bf \varepsilon}_t,
\]
where $f$ is a nonlinear function learned from data and ${\bf \varepsilon}_t$ is a noise term. We focus on models that admit an additive decomposition across source variables, enabling the interpretation of directed predictive relationships.

For each target variable $i$, the learned model induces a one-step-ahead predictive functional
\[
\hat{y}_{it} = F_i({\bf y}_{t-1}, \dots, {\bf y}_{t-p}),
\]
where $F_i(\cdot)$ is constructed as an additive composition of univariate component functions. The central object of interest in this work is not the function parameters themselves, but the dependence of $F_i$ on the lagged history of each source variable.

\textbf{Neural Additive Vector Autoregression (NAVAR)} (Bussmann et al., 2021) models each target variable $y_{it}$ as
\[
y_{it} = \sum_{j=1}^{N} \sum_{\ell=1}^{p} f_{ij\ell}(y_{j,t-\ell}) + \varepsilon_{it},
\]
where each $f_{ij\ell}(\cdot)$ is a learned univariate nonlinear function implemented as a shallow neural network. 
The additive structure of NAVAR implies that each source variable contributes independently to the prediction of $y_{it}$, conditional on the remaining variables. This design choice enables the isolation of individual lagged effects without requiring linearity or parametric assumptions about their form. In particular, the contribution of variable $j$ to target $i$ can be evaluated independently of other sources, which is essential for defining edge-level interventions. While the component functions $f_{ij\ell}$ are not interpreted as structural causal mechanisms, their additive separation ensures that removing a component corresponds to a well-defined modification of the predictive functional.

To encourage sparsity and stabilize estimation, the model is trained by minimizing mean squared error (MSE) with $\ell_1$ regularization applied to the \textit{outputs} of the contribution functions (not the network weights):
\[
{\cal L}
= \sum_{t,i} \left(y_{it} - \hat{y}_{it}\right)^2
+ \lambda \sum_{i,j,\ell} \| f_{ij\ell} \|_1,
\]
where $\|\cdot\|_1$ denotes the expected absolute output value of $f_{ij\ell}$ over its input support. This output-level penalty directly shrinks contribution magnitudes toward zero, encouraging a sparse causal graph. The model is trained using Adam optimizer with learning rate $3\times10^{-4}$, batch size 128, dropout rate 0.10, weight decay $10^{-3}$, $\lambda = 0.15$, $p = 8$ lags, and 600 epochs. All input series are $z$-score normalized prior to training. Full hyperparameters are reported in Table~1.

Following prior work (Bussmann et al., 2021), we summarize learned relationships using a causal score matrix 
$S \in {\bf R}^{N \times N}$, where each entry is defined as
\[
S_{ij}= \sum_{\ell=1}^{p}
{\rm Var}_t \left( f_{ij\ell}(y_{j,t-\ell}) \right).
\]
The variance operator in $S_{ij}$ captures the temporal variability of the learned contribution from variable $j$ to variable $i$. Intuitively, $S_{ij}$ measures how strongly and consistently changes in $y_{j,t-\ell}$ propagate through the learned function to affect predictions of $y_{it}$. As such, $S_{ij}$ summarizes contribution magnitude but does not encode whether this contribution is indispensable for prediction. 
 
Higher values of $S_{ij}$ indicate stronger and more consistent predictive contributions from variable $j$ to variable $i$. In systems with strong persistence or correlated predictors, large values of $S_{ij}$ may arise even when the corresponding contribution is redundant.

\begin{table}[t]
\centering
{\small
\caption{NAVAR hyperparameters used in all experiments.}
\begin{tabular}{lc}
\hline
{\bf Hyperparameter} & {\bf Value} \\
\hline
Maximum lag $p$ & 8 \\
Hidden layers per $f_{ij\ell}$ & 1 \\
Hidden units per layer & 32 \\
Activation & ReLU \\
Dropout rate & 0.10 \\
Weight decay ($\ell_2$) & $10^{-3}$ \\
Sparsity penalty $\lambda$ ($\ell_1$ on outputs) & 0.15 \\
Optimizer & Adam \\
Learning rate & $3\times10^{-4}$ \\
Batch size & 128 \\
Training epochs & 600 \\
Forecast loss & MSE \\
Input normalization & $z$-score \\
\hline
\end{tabular}
}
\end{table}

\subsection{Proposed Framework}

\textbf{Edge Ablation}. Formally, edge ablation corresponds to a restricted predictive functional
\[
F_i^{(-j)}({\bf y}_{t-1}, \dots, {\bf y}_{t-p})
= F_i({\bf y}_{t-1}, \dots, {\bf y}_{t-p})
\;\big|\; f_{ij\ell} \equiv 0 \;\; \forall \ell.
\]
To assess whether a candidate relationship $j\rightarrow i$ is necessary for prediction, we construct a restricted model in which all lagged inputs from variable $j$ are removed from the prediction of $y_i$:
\[
y_{it}^{(-j)}
= \sum_{k \neq j} \sum_{\ell=1}^{p}
f_{ik\ell}(y_{k,t-\ell}) + \varepsilon_{it}.
\]
This restriction defines a counterfactual model in which the information flow from variable $j$ to target $i$ is removed while all other predictive pathways are preserved. Unlike coefficient masking in linear models, this intervention removes an entire nonlinear contribution function, yielding a sharp test of necessity. Critically, the model is \textit{not retrained}: all component functions $f_{ik\ell}$ for $k \neq j$ remain exactly as learned, exploiting NAVAR's additive structure to isolate the marginal contribution of the candidate edge. 

All other aspects of training and evaluation are held fixed. This procedure isolates the marginal contribution of the candidate edge while preserving the remainder of the model structure.

\textbf{Forecast-Necessity Testing.} Let $L_t$ and $L_t^{(-j)}$ denote the out-of-sample MSE forecast losses at time $t$ for the full and masked models, respectively. We define the loss differential
\[
d_t = L_t^{(-j)} - L_t.
\]
The expected loss differential $E[d_t]$ can be interpreted as the increase in predictive risk incurred by removing the candidate relationship. A positive expected value indicates that the edge contributes uniquely to reducing forecast error, whereas a value near zero implies redundancy.

A positive mean of $d_t$ indicates that removing the edge degrades predictive performance. To test whether this degradation is systematic
rather than incidental, we apply a one-sided Diebold--Mariano test:
\[
H_0 : E[d_t] = 0
\quad {\rm vs.} \quad
H_1 : E[d_t] > 0.
\]
The DM test statistic is
\[
{\rm DM} = \frac{\bar{d}}{\sqrt{\widehat{\rm Var}_{\rm HAC}(\bar{d})}},
\]
where $\bar{d} = \frac{1}{T}\sum_t d_t$ is the mean loss differential and $\widehat{\rm Var}_{\rm HAC}$ denotes the heteroscedasticity and autocorrelation consistent (HAC) variance estimator, which accounts for serial correlation in the loss differential sequence, critical in time-series and panel settings (Diebold, 2015). 

Because the masked model is nested within the full model, the DM test may be slightly conservative in small samples; the HAC correction is adopted as a practical remedy and resulting $p$-values should be treated as approximate. Edges for which $H_0$ is rejected at significance level $\alpha = 0.05$ are classified as forecast-necessary for the target variable.

\textbf{Multiple Testing.} When testing all $N(N-1)$ directed edges simultaneously, the family-wise error rate is inflated. We report raw $p$-values for the focal Suffrage example in Table~2 and note that the result for Source~1 (Equal Protection) is robust even under Bonferroni correction ($p \approx 1.5 \times 10^{-8}$, well below $\alpha/N(N-1)$). We recommend Bonferroni or Benjamini-Hochberg correction when screening all edges as a discovery step.

\textbf{Interpretation Outputs.} Taken together, the causal score matrix and forecast-necessity tests define a two-stage interpretation pipeline. Causal scores identify candidate relationships based on average contribution, while necessity tests determine which of these relationships are required for accurate prediction. Disagreement between the two signals potential redundancy or substitutability.

The proposed framework yields three complementary outputs:

\begin{itemize}
\item Causal scores $S_{ij}$, summarizing average predictive contribution.
\item Necessity indicators, identifying relationships required for accurate forecasting.
\item Local explanations, used diagnostically to understand how necessary relationships operate over time.
\end{itemize}
Crucially, forecast necessity is evaluated independently of score magnitude, enabling principled distinctions between frequent, large, and indispensable effects.

\section{Case Study: Interpreting Causal Relationships in Democratic Development}

\subsection{Motivation and Scope}
We illustrate the proposed forecast-necessity framework using a case study of democratic development, modeled as a multivariate time series of democracy indicators observed across countries. This domain provides a useful testbed for evaluating causal reasoning in nonlinear time-series models for three reasons. First, democratic indicators exhibit strong temporal persistence, making it easy for models to over-attribute importance to inertia rather than cross-variable influence. Second, many indicators are conceptually related and empirically correlated, creating redundancy and substitutability among predictors. Third, causal interpretations in this domain are substantively consequential, highlighting the value of distinguishing apparent importance from genuine predictive necessity (Coppedge et al., 2020; Coppedge et al., 2022).

The goal of this analysis is not to establish structural causal effects in democratic development. Rather, it is to demonstrate how different notions of causal relevance - magnitude-based versus forecast-based - lead to qualitatively different interpretations in a real-world, interpretable system.

\subsection{Data}

To empirically evaluate the proposed framework, we draw on data from the Varieties of Democracy (V-Dem) project, one of the most comprehensive and widely used datasets for measuring democratic institutions and governance (Coppedge et al., 2020). V-Dem provides expert-coded, theoretically grounded indicators capturing multiple dimensions of democracy across countries and time.

We use version 15 of the V-Dem country-year dataset, which covers the period from 1789 to 2024 (Coppedge et al. 2025). From this corpus, we select 16 lower-level democracy components that serve as foundational building blocks for V-Dem's higher-order democracy indices. These components are theoretically well-motivated, empirically validated, and commonly used in the construction of aggregate democracy measures, making them particularly suitable for causal analysis of democratic dynamics. The dataset consists of annual indicators measuring dimensions such as freedom of expression, freedom of association, suffrage, electoral integrity, judicial constraints, and legislative constraints, among others, observed for multiple countries over time. Each country is treated as an independent time series, and the data are modeled as a panel with relatively short temporal length per unit.

To ensure suitability for nonlinear time-series modeling and resampling-based inference, we apply several restrictions to the raw data. First, we focus on the period 1990-2024, excluding earlier years. This temporal restriction is substantively and methodologically motivated. Substantively, a large number of contemporary states emerged only after the dissolution of multinational entities such as the Soviet Union and Yugoslavia, and therefore do not have meaningful observations prior to 1990. Methodologically, restricting the sample reduces structural breaks associated with decolonization and early state formation, yielding more comparable time-series dynamics across units.

Second, we retain only countries with balanced and complete observations on all selected democracy components over the full analysis window. Countries with substantial missingness or discontinuous coverage are excluded to avoid imputation-driven artifacts and to ensure valid block-bootstrap resampling and out-of-sample forecasting.

The final dataset consists of 139 countries, each observed for 35 consecutive years (1990-2024), yielding a strongly balanced panel suitable for causal time-series modeling. This structure allows us to estimate nonlinear interdependencies among democratic components while preserving temporal dependence, cross-variable feedback, and comparability across units.\\

\subsection{Training and Evaluation Protocol}

For each country, the first 30 years (1990-2019) are used for NAVAR training and the final 5 years (2020-2024) are held out as the validation set for out-of-sample forecast evaluation and DM testing. Lag windows are constructed so that they never cross country boundaries. Hyperparameters are as reported in Table~1.

\subsection{Panel Structure and Evaluation}

The democratic development data analyzed in this work exhibit a panel structure, consisting of multiple countries observed over time. Let $c = 1, \dots, C$ index countries, and denote the corresponding country-specific time series by
\[
{\bf y}_{c,t}, \quad t = 1, \dots, T_c.
\]

Each country is treated as an independent realization of the same underlying dynamic system, sharing a common predictive functional $F_i(\cdot)$. Model estimation is performed on the pooled panel data, allowing information to be shared across units while preserving within-country temporal dependence.

Forecast evaluation and necessity testing are conducted out of sample and respect the panel structure. Let
\[
L_{c,t}
\quad \rm{and} \quad
L^{(-j)}_{c,t}
\]
denote the forecast losses for the full and restricted models, respectively, at country $c$ and time $t$. We define the country-level loss differential as
\[
d_{c,t} = L^{(-j)}_{c,t} - L_{c,t}.
\]

The Diebold-Mariano test is applied to the pooled sequence $\{d_{c,t}\}$, accounting for serial dependence over time within units. This procedure evaluates whether a candidate relationship is systematically necessary for prediction across the panel, rather than being driven by idiosyncratic dynamics in a small number of countries.

Importantly, this treatment does not assume homogeneous causal effects across units. Instead, it tests whether a predictive relationship is generically necessary for accurate forecasting in the system as a whole.

\subsection{Modeling}
We train a Neural Additive Vector Autoregressive (NAVAR) model on the pooled panel data using a fixed lag window and $\ell_1$ regularization. The model induces a directed graph over variables, summarized by a causal score matrix as described in previous section. Throughout the case study, this directed variable-level graph is treated as the primary object of interpretation.

\subsection{Global Causal Structure from NAVAR Scores} 
\begin{figure}[ht]
\vskip 0.2in
\begin{center}
\centerline{\includegraphics[width=\columnwidth]{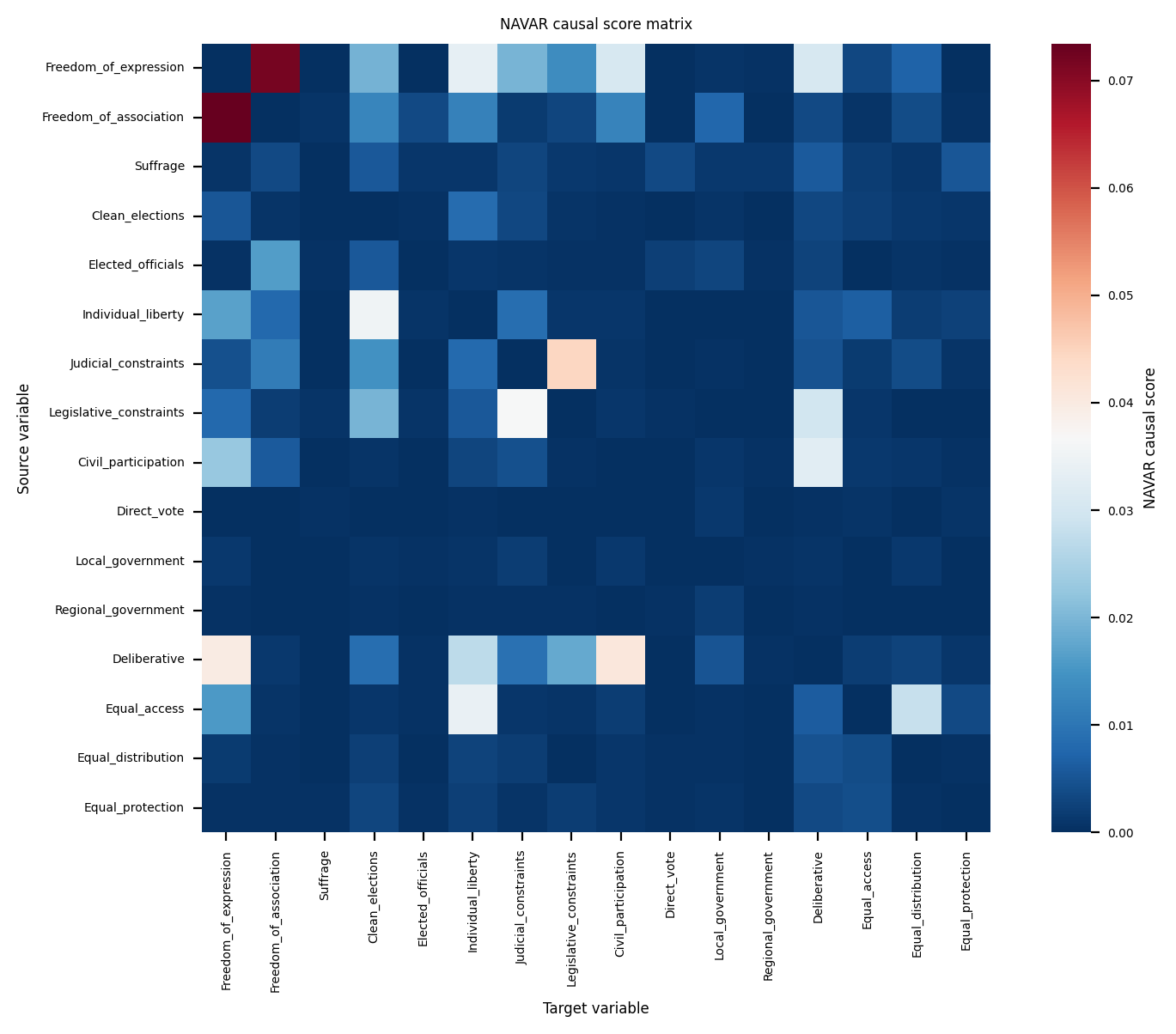}}
\alt{Heatmap of the NAVAR causal score matrix for 16 democracy variables.
Diagonal entries are zeroed out. Darker blue cells indicate higher causal
scores; a few off-diagonal entries show elevated scores for variables such
as Freedom of Expression and Freedom of Association.}
\caption{
NAVAR Causal Score Matrix computed on the democracy panel. Each
entry is the temporal standard deviation of the learned contribution from
source $i$ to target $j$. Diagonal entries (self-effects) are zeroed out.
}
\label{fig:navar_matrix}
\end{center}
\end{figure}

Figure~\ref{fig:navar_matrix} presents the NAVAR causal score matrix for the democratic development system. Diagonal entries dominate the matrix, reflecting strong temporal persistence in all indicators, and for ease of interpretation, are removed from the figure. Several cross-variable relationships exhibit high scores, suggesting substantial predictive influence under a magnitude-based interpretation.

At this stage, a naive reading might conclude that variables with larger causal scores are more causally important. However, this interpretation does not distinguish between persistent self-dependence, redundant predictors, and relationships that are genuinely required for prediction.

\subsection{Forecast-Necessity Testing via Edge Ablation}
To assess whether high-scoring relationships are actually necessary for prediction, we apply the proposed edge-ablation procedure to NAVAR models. For each candidate edge, we compare out-of-sample forecast accuracy between the full NAVAR model and a restricted model in which the source variable's lagged history is removed from the target equation.

Table~\ref{tab:DM_test} reports forecast-necessity results for ``Suffrage'' as the target variable, comparing two candidate source variables using NAVAR causal scores and Diebold-Mariano (DM) forecast-comparison tests.

Under a magnitude-based interpretation, Source 2 (``Equal Access'') would be judged more important: it exhibits the higher NAVAR causal score ($0.0065$), which is substantively meaningful for scores of this scale in the democratic development system. By contrast, Source 1 (``Equal Protection'') has a slightly lower causal score ($0.0056$). If causal relevance were inferred from score magnitude alone, Source 2 would therefore be ranked as the more influential predictor of suffrage.

The forecast-necessity results contradict this interpretation. Removing Source 1 from the model increases the mean forecast loss by one order of magnitude relative to Source 2, producing a large positive loss differential and a highly significant DM statistic (DM = $5.54$, $p \approx 1.5 \times 10^{-8}$). This indicates that Source 1 is forecast-necessary for predicting suffrage: its contribution cannot be recovered once removed.

In contrast, removing Source 2, despite its higher NAVAR causal score, produces virtually no change in predictive accuracy. The mean loss remains essentially unchanged, and the DM test fails to reject the null hypothesis (DM = $1.02$, $p = 0.154$). Source 2 is therefore not forecast-necessary, even though its average contribution is large enough to appear important under a score-based interpretation.

These results illustrate a central finding: causal score magnitude and forecast necessity are not equivalent. Some relationships contribute frequently but weakly and are indispensable for prediction, while others contribute intermittently or redundantly and can be substituted by correlated predictors.

\begin{table}[htbp]
\centering
{\small
\caption{Results of Diebold-Mariano (DM) Forecast-Necessity Test. Loss is MSE evaluated on the held-out validation panel (2020-2024). The masked model zeros out the source variable's contribution at evaluation time without retraining.}
\label{tab:DM_test}
\begin{tabular}{lcc}
\hline
{\bf Target} & {\bf Source 1} & {\bf Source 2} \\
Suffrage & Equal Protection & Equal Access \\
\hline
{\bf Measure} & {\bf Source 1} & {\bf Source 2} \\
NAVAR Score & 0.0056342 & 0.0065086\\
Mean Loss Full & 0.5766956 & 0.5766956\\
Mean Loss Masked & 0.5767076 & 0.5766984 \\
Mean Diff. & 0.0000176 & 0.0000026 \\
DM stat & 5.5377072 & 1.0179710 \\
$p$-value & 0.0000000 & 0.1543458 \\
Forecast-necessary? & {\bf Yes} & No \\
\hline
\end{tabular}
}
\end{table}

\subsection{Local Explanations and Functional Effects} 
\begin{figure}[ht]
\vskip 0.2in
\begin{center}
\centerline{\includegraphics[width=\columnwidth]{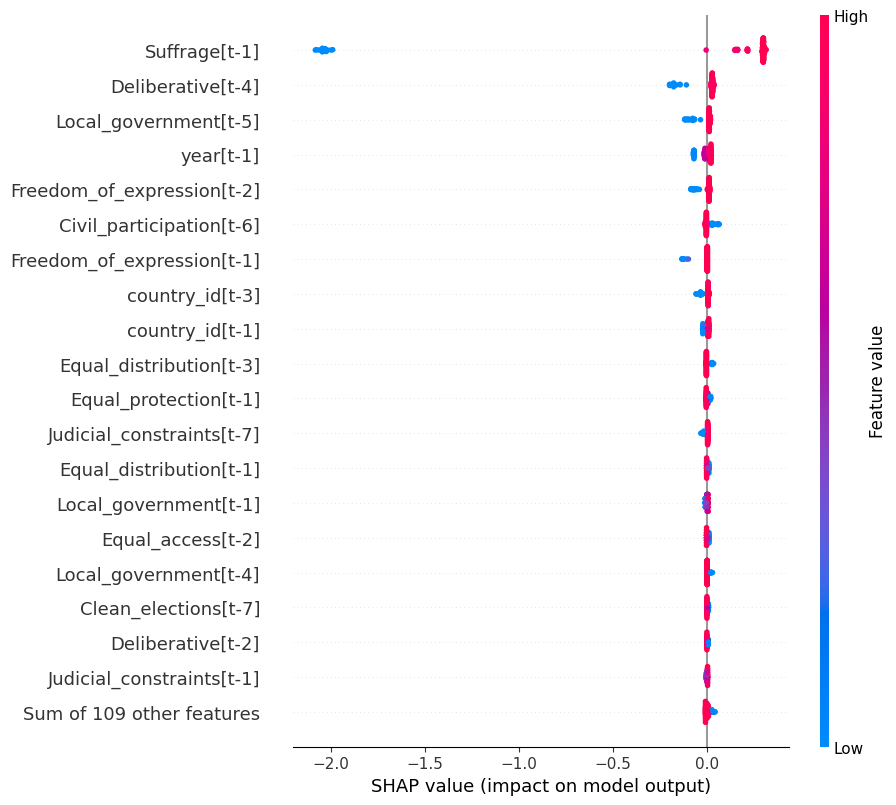}}
\alt{SHAP beeswarm plot for the Suffrage variable. The largest feature
impact comes from Suffrage at lag 1 (self-persistence). Equal Protection
at lag 1 shows a consistent negative influence, while Equal Access at lag
2 shows a smaller, less consistent contribution.}
\caption{
SHAP-based local explanations for the {\em Suffrage} target.
Equal Protection (Source~1) exhibits a consistent lag-1 influence, while
Equal Access (Source~2) contributes at lag~2 with lower amplitude,
explaining why Source~1 is forecast-necessary but Source~2 is not.
}
\label{fig:shap}
\end{center}
\end{figure}

To further understand why certain relationships are necessary, we examine local explanations and functional contribution plots produced by the NAVAR model. Figure~\ref{fig:shap} shows SHAP-based summaries for predicting Suffrage at representative time points. Predictions are dominated by short-term persistence, with smaller but consistent contributions from different variables. It also confirms the results we've obtained with the DM test: that Source 1 (``Equal Protection'') has a more important influence with a lag $t_1$, whereas Source 2 (``Equal Access'') has less influence, and moreover, with a lag of $t-2$.

These diagnostics clarify how necessary relationships operate over time, complementing the binary necessity results obtained from forecast comparison.

\subsection{Lessons for Interpretable Causal Reasoning}

This case study highlights several lessons for interpreting causal outputs from nonlinear time-series models. First, temporal persistence can dominate magnitude-based scores, causing models to attribute importance to inertia rather than to cross-variable influence. In highly persistent systems, large causal scores may therefore reflect stability over time rather than substantive interaction among variables.

Second, small but stable effects may be more important for prediction than larger, sporadic contributions. Relationships that operate consistently across time, even with modest average magnitude, can be indispensable for forecasting, while relationships with large but irregular effects may be redundant or easily substituted.

Third, redundancy among predictors can inflate causal scores without implying necessity. When multiple variables encode overlapping information, removing one may have little impact on predictive performance despite its apparent importance under a score-based interpretation.

Most importantly, evaluating causal relevance through forecast necessity shifts interpretation from parameter inspection to model behavior. By grounding causal claims in observable changes in predictive accuracy, forecast-necessity testing yields interpretations that are more robust, transparent, and easier to justify than coefficient-centric analogues in nonlinear time-series models.

\section{Discussion}
This work highlights a fundamental limitation of magnitude-based interpretations in nonlinear time-series models. Causal scores derived from neural autoregressive systems summarize average predictive contributions, but they do not indicate whether a relationship is required for accurate prediction. As demonstrated in the case study, relying on magnitude alone can lead to systematic misinterpretation, particularly in persistent and redundant systems.

Forecast-necessity testing addresses this gap by shifting attention from parameter values to model behavior. Evaluating whether predictive performance degrades when a relationship is removed yields a criterion for causal relevance that is operational, testable, and aligned with counterfactual intuition. Importantly, this approach does not require structural assumptions or domain-specific causal theory, making it broadly applicable across domains.

\subsection{Relationship to Existing Interpretability Tools} 
Forecast-necessity testing complements, rather than replaces, existing interpretability methods. Feature-attribution techniques help explain how predictions are formed locally, while necessity tests evaluate whether those contributions are indispensable. Together, these tools provide a more complete picture of causal reasoning in complex models.

Unlike coefficient-based significance tests, forecast-necessity testing remains meaningful in nonlinear, regularized settings and avoids overinterpreting artifacts of persistence or collinearity. This makes it especially suitable for applied AI systems where interpretability errors carry real consequences.

\subsection{Limitations and Scope} The proposed framework evaluates causal relevance relative to a specific predictive model and forecasting task. First, necessity is model- and context-dependent, and should not be conflated with structural causality. Second, the evaluation is demonstrated on a single modeling framework (NAVAR) and a single domain (democracy indicators), which limits claims about generalizability; extending the approach to other nonlinear architectures and domains is an important direction for future work. Third, the DM test applied to nested models may be slightly conservative; future work could adopt the Clark-West correction, which could be better calibrated for nested forecast comparisons. Fourth, when testing all $N(N-1)$ edges simultaneously, multiple-testing corrections are necessary and are currently left to the analyst. Fifth, the ablation procedure masks contributions at evaluation time without retraining; this is efficient and appropriate given NAVAR's additive structure, but does not account for how remaining components might adapt if the model were retrained without the ablated variable. Finally, the approach does not scale trivially to very high-dimensional settings, where the number of edge tests grows quadratically in the number of variables $O(N^2)$.

Despite these limitations, forecast-necessity testing provides a principled and practical diagnostic for interpreting causal outputs where conventional statistical inference is unavailable or inappropriate.

\section{Conclusion}
Interpreting causal relationships in nonlinear time-series models requires more than inspecting large coefficients or scores. In this paper, we introduced forecast necessity as an alternative criterion for causal relevance and presented a simple evaluation framework based on edge ablation and forecast comparison.

Using Neural Additive Vector Autoregression and a case study of democratic development, we showed that causal score magnitude and predictive necessity can diverge substantially, with important implications for how model outputs are interpreted. Forecast-necessity testing yields clearer, more defensible explanations by identifying which relationships are required for accurate prediction rather than merely prominent on average.

More broadly, this work argues for prediction-centric reasoning in applied causal discovery. By grounding causal interpretation in observable changes in model performance, forecast-necessity testing supports more reliable and transparent reasoning in complex dynamic systems. Looking ahead, the forecast-necessity criterion introduced here has been extended in subsequent work to serve as a structural prior for dynamic causal inference (Kuskova, Zaytsev, and Coppedge, 2026), where learned necessity masks constrain time-varying network autoregression and enable impulse-response 
analysis under structural uncertainty. In addition, adopting nested-model forecast comparison tests such as Clark-West would improve inferential calibration. Extending the framework to non-additive architectures would broaden its applicability. Connecting forecast necessity to sensitivity analysis or partial identification could help bridge the gap between predictive and structural causal reasoning - an important open problem as neural causal discovery is deployed in increasingly high-stakes scientific and policy domains.

\section{References} 
Albini, E., Long, J., Dervovic, D., \& Magazzeni, D. (2022). Counterfactual shapley additive explanations. In Proceedings of the 2022 ACM conference on fairness, accountability, and transparency (pp. 1054--1070).

Aoki, M. (2013). State space modeling of time series. Springer Science \& Business Media.

Bussmann, B., Nys, J., and Latr\'{e}, S. Neural additive vector autoregression models for causal discovery in time series. In International Conference on Discovery Science, pp. 446--460. Cham: Springer International Publishing, 2021.

Chen, X., Chen, Y., Shen, Z., and Xiu, D. Recurrent neural networks for nonlinear time series, 2025.

Coppedge, M., Gerring, J., Glynn, A., Knutsen, C. H., Lindberg, S. I., Pemstein, D., and Marquardt, K. Varieties of democracy: Measuring two centuries of political change. Cambridge University Press, 2nd edition, 2020.

Coppedge, M., Edgell, A. B., Knutsen, C. H., and Lindberg, S. I. Why democracies develop and decline. Cambridge University Press, 2022.

Coppedge, M., Gerring, J., Knutsen, C.H., McMann, K., Mechkova, V., Medzihorsky, J., Natsika, N., Neundorf, A., Paxton, P., Pemstein, D., von R\"{o}mer, J., Seim, B., Sigman, R., Skaaning, S.-E., Staton, J., Sundstr\"{o}m, A., Tannenberg, M., Tzelgov, E., Wang, Y.-T., Wig, T., Wilson, S. and Ziblatt, D. (2025) V-Dem Country-Year Dataset v15. Varieties of Democracy (V-Dem) Project. \url{https://www.v-dem.net/data/the-v-dem-dataset/} (Last accessed: 1/21/2026).

Diebold, F. X. (2015). Comparing predictive accuracy, twenty years later: A personal perspective on the use and abuse of Diebold--Mariano tests. Journal of Business \& Economic Statistics, 33(1), 1--9.

Geiger, P., Zhang, K., Schoelkopf, B., Gong, M., and Janzing, D. Causal inference by identification of vector autoregressive processes with hidden components. In International Conference on Machine Learning, pp. 1917--1925. PMLR, 2015.

Job, S., Tao, X., Cai, T., Xie, H., Li, L., Li, Q., \& Yong, J. (2025). Exploring Causal Learning Through Graph Neural Networks: An In-Depth Review. Wiley Interdisciplinary Reviews: Data Mining and Knowledge Discovery, 15(2), e70024.

Kment, B. (2006). Counterfactuals and the Analysis of Necessity. Philosophical Perspectives, 20, 237--302.

Kruschel, S., Hambauer, N., Weinzierl, S., Zilker, S., Kraus, M., \& Zschech, P. (2025). Challenging the performance-interpretability trade-off: an evaluation of interpretable machine learning models. Business \& Information Systems Engineering, 1--25.

Kuskova, V., Zaytsev, D., and Coppedge, M. (2026). From causal  discovery to dynamic causal inference in neural time series. arXiv preprint arXiv:2603.20980.

Kwon, H. J., Koo, H. I., \& Cho, N. I. (2021). Improving explainability of integrated gradients with guided non-linearity. In 2020 25th International Conference on Pattern Recognition (ICPR) (pp. 385--391). IEEE.

Li, Y. and Genton, M. G. Single-index additive vector autoregressive time series models. Scandinavian Journal of Statistics, 36(3):369--388, 2009.

Lim, N., d'Alch\'{e}-Buc, F., Auliac, C., \& Michailidis, G. (2015). Operator-valued kernel-based vector autoregressive models for network inference. Machine Learning, 99(3), 489--513.

Mehdiyev, N., Enke, D., Fettke, P., \& Loos, P. (2016). Evaluating forecasting methods by considering different accuracy measures. Procedia Computer Science, 95, 264--271.

Montagna, F., Noceti, N., Rosasco, L., Zhang, K., and Locatello, F. Scalable causal discovery with score matching. In Conference on Causal Learning and Reasoning, pp. 752--771. PMLR, 2023.

Nichols, J. D., \& Cooch, E. G. (2025). Predictive models are indeed useful for causal inference. Ecology, 106(1), e4517.

Taskaya-Temizel, T., and Casey, M. C. (2005). A comparative study of autoregressive neural network hybrids. Neural Networks, 18(5--6), 781--789.

Tramontano, D., Kivva, Y., Salehkaleybar, S., Drton, M., and Kiyavash, N. Causal effect identification in LiNGAM models with latent confounders. In Proceedings of the 41st International Conference on Machine Learning, pp. 48468--48493. PMLR, 2024.

Troster, V., and Wied, D. A specification test for dynamic conditional distribution models with function-valued parameters. Nonlinear Dynamics, 2:109--127, 2021.

Varieties of Democracy Dataset. \url{https://www.v-dem.net/}

Wan, H., Wang, H., Gu, C., and Yang, H. Causality structures in nonlinear dynamical systems. Nonlinear Dynamics, 113(10):11455--11475, 2025.

Zhang, W., Panum, T., Jha, S., Chalasani, P., and Page, D. Cause: Learning granger causality from event sequences using attribution methods. In International Conference on Machine Learning, pp. 11235--11245. PMLR, 2024.

Zhou, W., Bai, S., Yu, S., Zhao, Q., and Chen, B. Jacobian regularizer-based neural granger causality. In Proceedings of the 41st International Conference on Machine Learning, pp. 61763--61782. PMLR, 2024.
\bigskip
\end{document}